# Lightweight LiDAR-Based Cone Detection Framework Using Random Forest for Formula Student Driverless

Márk Mező-Kerekes[1[0009-0001-7878-1957]], Péter Praksz [1[0009-0004-7080-3252]]

and Chang Liu[1,2 [0000-0001-6610-5348]]

[1] Department of Networked Systems and Services, Faculty of Electrical Engineering and Informatics, Budapest University of Technology and Economics, Műegyetem rkp. 3, H-1111 Budapest, Hungary

[2] Machine Perception Research Laboratory, HUN-REN Institute for Computer Science and Control (SZTAKI), Kende u. 13–17, H-1111 Budapest, Hungary

mezo-kerekes.mark@edu.bme.hu praksz.peter@edu.bme.hu
changliu@hit.bme.hu

**Abstract.** Reliable, low-latency perception is crucial for Formula Student Driverless vehicles, yet many existing pipelines rely on deep learning and multi-sensor fusion, often requiring GPU acceleration. This paper presents a lightweight LiDAR-only perception pipeline tailored for CPU execution, combining ground removal, IMU-based motion compensation, DBSCAN clustering, and geometric feature-based Random Forest classification. Feature importance analysis reduced the model input from 12 to 7 features while preserving performance. Evaluated on 2,371 labeled clusters collected from real FSD events, the pipeline achieves an F1-score of 98.33% and an end-to-end runtime of 3.13 ms on CPU-only hardware. The released dataset, labeling tool, and trained models provide a practical and reproducible baseline for other resource-constrained autonomous racing teams.

**Keywords:** LiDAR, Random Forest, Formula Student, Real-Time Perception, Autonomous Racing, CPU Inference

## 1 Introduction

The autonomous vehicle market is projected to reach $4.45 trillion by 2034, growing at an annual rate of 36.3% annually [1]. This rapid development highlights the growing demand for reliable and computationally efficient perception systems capable of operating under strict latency and hardware constraints.

The Formula Student Driverless (FSD) competition emerged from the popularity of autonomous driving as a testbed for university students to put theory into practice and race against each other. Teams design and build fully autonomous race cars capable of navigating previously unknown tracks using only onboard sensors. Track boundaries are defined exclusively by traffic cones.

Deploying perception systems in FSD environments poses significant challenges. These systems must simultaneously operate under strict computational budgets, compensate for dynamic vehicle motion, and deliver accurate cone detection, all while relying on limited annotated datasets. Hardware constraints represent an unavoidable limitation: while well-funded teams may access high-end computing resources, many student teams cannot afford top-of-the-line hardware without compromising other aspects of their vehicle, making GPU-dependent solutions infeasible. Existing approaches either require GPU acceleration, suffer from reduced detection accuracy at distance, or depend on multi-sensor fusion architectures, leaving a gap for lightweight LiDAR-only perception methods optimized for CPU-based platforms.

In this paper we present a LiDAR-based perception pipeline tailored for the FSD environment. Compared with more complex deep learning approaches, the proposed method operates efficiently on CPU-based systems without GPU acceleration.

The main contributions of this paper are:

- Importance-based feature reduction, which improves computational efficiency while maintaining classification performance.
- A public labeling and training tool for labeling LiDAR cone clusters and training models on the labeled data.
- A manually annotated dataset of 2,371 labeled clusters collected from real FSD events including yellow, blue, and orange cones.
- Reproducible trained models that can be used as a baseline for future improvements.

By presenting this approach, we aim to provide a practical and publicly accessible perception solution for FSD teams, while also demonstrating the potential of light-weight machine learning methods for real-world autonomous systems.

## 2 Related Work

Related work in autonomous driving perception falls into four groups: multi-sensor fusion, existing FSD solutions, LiDAR-only deep learning and traditional machine learning [17].

Multi-sensor fusion approaches combine LiDAR, camera, and radar data to achieve robust detection under diverse conditions [4,5]. Several FSD teams follow this strategy [6,7,8], achieving reliable perception at the cost of system complexity and hardware accessibility. While accurate, these methods require substantial computational power and annotated datasets, limiting applicability in resource-constrained platforms. To address the scarcity of FSD-specific training data, the community-driven coneScenes [14] dataset was developed.

LiDAR-only deep learning methods such as PointPillars [2] and TimePillars [3] reduce sensor complexity but still depend on large datasets and GPU acceleration. Rule-based pipelines combining Euclidean or Density-Based Spatial Clustering of Applications with Noise (DBSCAN) clustering with geometric filtering have demonstrated viable FSD results [15], yet require careful manual tuning.

Traditional machine learning methods using handcrafted geometric features and classifiers, such as Support Vector Machines and Random Forests, offer a lightweight alternative with a favorable efficiency-accuracy trade-off for CPU-based real-time systems, yet lightweight, LiDAR-only methods optimized for CPU-based FSD environments remain underexplored. A gap this work directly addresses.

# 3 Methodology

The proposed perception pipeline is designed to achieve accurate cone detection under strict computational constraints, enabling real-time performance on CPU-based systems. The pipeline processes 20 Hz LiDAR scans acquired from an Ouster OS1-32 sensor mounted on the vehicle. As illustrated in Fig. 1, the pipeline consists of two main stages: preprocessing and Random Forest classification.

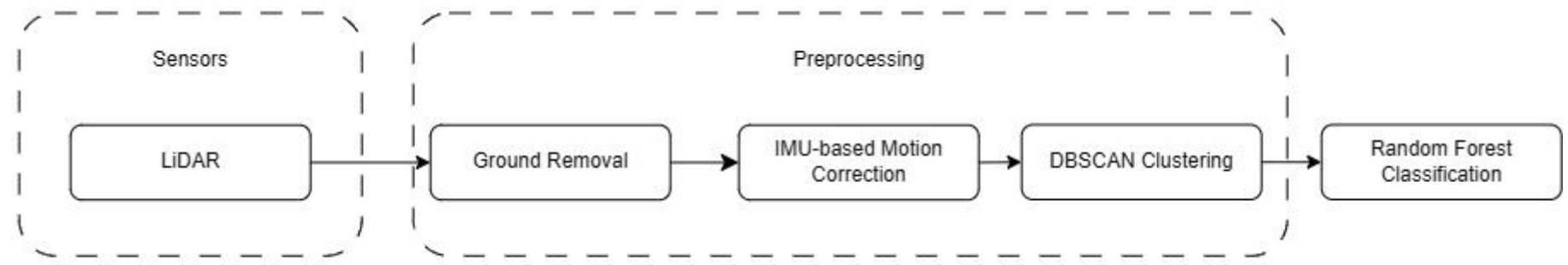

**Fig. 1. LiDAR perception pipeline**

## 3.1 Dataset

The dataset used in this study is custom-built and manually annotated. It consists of clustered LiDAR segments extracted from real FSD scenarios, including Skidpad, Acceleration, and Autocross tracks, ensuring diversity in track layouts and environmental conditions. To support efficient and consistent annotation, a custom open-source labeling tool, Cone-Cluster-Labeler[16], was developed. The tool enables visualization and labeling of clustered LiDAR data within a ROS2-based framework, significantly improving annotation efficiency and consistency.

The final dataset comprises 2,371 samples, including 1,164 cone and 1,207 non-cone instances, resulting in a balanced class distribution. Although relatively small, it is suitable for training models suited for low-data regimes such as Random Forest. To reduce temporal correlation, only every $10^{th}$ LiDAR frame was retained prior to annotation. To prevent train/test split information leakage, clusters originating from the same run were kept together and were not split across the training and test subsets. Consequently, the class distribution differs between the two sets: the test split contains approximately 71.8% cone and 28.2% non-cone samples, whereas the training split contains approximately 42.9% cone and 57.1% non-cone samples. The dataset was partitioned into 1,867 training samples and 504 held-out test samples.

### 3.2 Preprocessing

Due to the angled mounting of the LiDAR sensor, a large portion of the captured point cloud corresponds to the ground surface. To reduce computational overhead and improve downstream efficiency, a ray-based ground removal algorithm is applied. For each point, polar coordinates are computed and a sector index is assigned:

$$S_i = \left\lfloor \frac{\theta_i}{\Delta\theta} \right\rfloor \tag{1}$$

where $S_i$ is the sector index, $\theta_i$ is the azimuth angle of point $i$, and $\Delta\theta$is the angular width of each sector. This sector-based representation enables efficient local processing and reduces computational complexity.

Within each sector, points are sorted by radial distance, and both global and local height thresholds are computed, where $\Delta r$ is the distance between $P_i$ and previous $P_{i-1}$ point. $|\Delta z|$ is the maximum allowed global height threshold, while $|z_i|$ is the local maximum height in each sector. A point is classified as ground if both local and global constraints are satisfied:

$$|\Delta z| \leq h_{global}, \; |z_i| \leq h_{local} \tag{2}$$

where,

$$h_{global} = \tan(\beta_{max}) * r_i \tag{3}$$

$$h_{local} = \min(\tan(\alpha_{max}) * \Delta r, \; h_{global}) \tag{4}$$

This method was inspired and was slightly modified for our purposes by these [9,10] articles.

In high-speed driving conditions, motion distortion occurs due to the sequential acquisition of LiDAR points within a scan period of ~50 ms. This distortion can negatively impact precise cone localization. To mitigate this effect, motion compensation is performed using IMU measurements. For each LiDAR $P_i$, the corrected position is:

$$P = R * P_i + \Delta\vec{d} \tag{5}$$

where $R$ is the rotational matrix, $\Delta\vec{d}$ is the linear displacement. The rotation is approximated as:

$$R = I + \sin(\theta)\, W + (1 - \cos(\theta))W^2 \tag{6}$$

where $W$ is the skew-symmetric matrix of the angular velocities from the IMU.

Following the motion correction, the remaining point cloud is divided into clusters using DBSCAN [13], which groups spatially dense regions, while effectively handling noise without requiring a predefined number of clusters.

### 3.3 Feature Extraction

From each cluster, a set of geometric features is extracted to characterize the spatial and structural properties of candidate objects. Initially 12 geometric features were considered. Feature selection was performed based on importance scores derived from the Random Forest model, allowing identification and removal of redundant features. Features contributing less than 1% to overall model importance were removed, resulting in a reduced set of seven features. This reduction improves computational efficiency, while preserving and slightly improving classification performance. The selected features are summarized in Table 1 and are designed to capture both the geometric shape and spatial distribution of clustered objects.

| Name | Formula | Comments |
|---|---|---|
| Height | $\max(P_z) - \min(P_z)$ | $P_z$ is the Z -coordinate of the cluster's points |
| Width | $\max(\Delta P_x, \Delta P_y)$ | $\Delta P_x, \Delta P_y$ denote the ranges of the cluster along the x- and y-axes |
| Depth | $\min(\Delta P_x, \Delta P_y)$ | |
| Aspect Ratio | $\frac{Height}{Width}$ | |
| Volume | $Height * Width * Depth$ | |
| Density | $\frac{N}{Volume}$ | $N$ denotes the number of points in the cluster |
| Distance from LiDAR | $\sqrt{c_x^2 + c_y^2 + c_z^2}$ | $c$ denotes the centroid of the cluster |

**Table 1.** Extracted geometric features

To ensure consistency across different viewpoints, width and depth are defined dynamically: the larger spread between x- and y-axes is assigned as width, while the smaller is treated as depth. This normalization improves robustness to orientation variations and resulted in an approximate 1% increase in F1-score.

### 3.4 Classification

The extracted features are used as input to a Random Forest classifier [12]. Each cluster is treated as a candidate object and classified as either cone or non-cone. Random Forest is selected due to its favorable trade-off between computational efficiency and classification performance. In contrast to deep learning-based methods, which typically require large-scale datasets and GPU acceleration, Random Forest can be effectively trained on small datasets and executed efficiently on CPU-only systems. Additionally, the model demonstrates robustness against overfitting in low-data regimes, which is critical in Formula Student environments where large annotated datasets such as NuScenes [11] are not directly applicable. Hyperparameters were optimized using grid search with 5-fold cross-validation. The final model consists of 65 estimators with a maximum depth of 10, with a fixed random seed (42) to ensure reproducibility.

## 4 Results

The proposed pipeline was evaluated on the held-out test split (504 samples) of the custom dataset. All runtime measurements were performed exclusively on an Intel Core Ultra 9 285H processor with 32GB RAM, running Ubuntu 24.04 in performance mode, while operating on battery power.

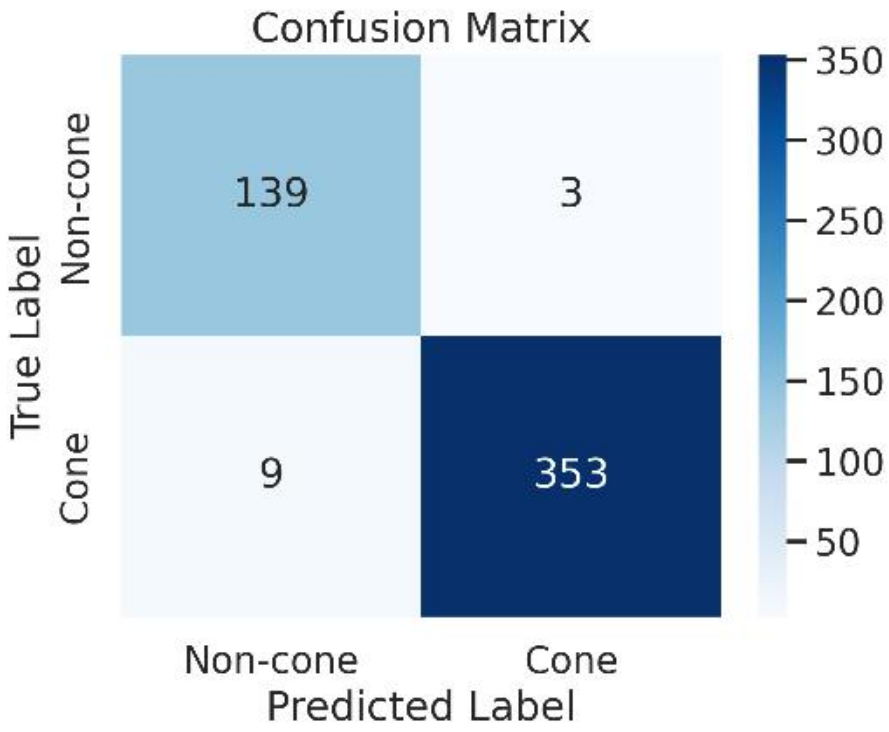

**Fig. 2.** Confusion Matrix on Test Split

The Random Forest classifier achieved an F1-score of 98.33%. In the FSD context false negatives are more critical, since missing a cone could cause the vehicle to leave the track. However, this risk is substantially mitigated by the pipeline's low latency as consecutive frames are unlikely to produce consistent misclassification.

The complete perception pipeline achieved an end-to-end runtime of 3.13 ms, corresponding to a maximum theoretical throughput of approximately 317 Hz, which is significantly higher than the 20 Hz LiDAR scan rate.

| Competitor Name | Runtime |
|---|---|
| AMZ Driverless | 250ms |
| FST | 21.4ms |
| Delft | 90ms |
| **BME FRT (Ours)** | **3.13ms** |

**Table 2.** Runtime Comparison with Published FSD Systems

Published FSD perception systems generally trade runtime for accuracy. AMZ Driverless achieves near-100% detection accuracy, but reports performance degradation beyond 4.5 m [6]. The FST pipeline achieves comparable cone detection in 13.77 ms on CPU hardware, while additionally incorporating color classification [7]. The Delft system achieves 98% test set accuracy up to 6 m [8]. The proposed pipeline maintains 98.33% F1-score up to 15m, the practical detection horizon used during evaluation, while achieving lower end-to-end runtime than each of the systems and operating without GPU. Direct runtime comparisons are inherently approximate, as competing teams report timings on different hardware.

## 5 Discussion, Limitations and Conclusion

The results confirm that a compact set of geometric features derived from physical properties combined with a Random Forest classifier is sufficient for FSD cone detection tasks without GPU or large training data. The 3.13 ms runtime leaves ample processing budget for downstream path planning and control. As shown in Table 3, feature reduction consistently improves end-to-end runtime while precision and recall remain stable, indicating that the removed features were computationally redundant.

| Feature Number | Precision | Recall | F1-Score | Runtime |
|---|---|---|---|---|
| 12 | 98.87% | 96.69% | 97.77% | † |
| 9 | 99.15% | 96.96% | 98.04% | 4.54ms |
| 8 | 99.15% | 96.69% | 97.90% | 3.50ms |
| 7 | 99.16% | 97.51% | 98.33% | 3.13ms |

†Runtime not measured

**Table 3.** Performance Metrics and end-to-end Runtime for Reduced Feature Sets

More broadly, this work demonstrates that traditional machine learning remains competitive with deep learning in resource-constrained real-time perception tasks. With an average RAM usage of approximately 121MB, the pipeline is suitable for deployment on resource-constrained hardware without modification.

The Random Forest classifier provides class probability estimates, which can be integrated into downstream mapping and path planning modules for uncertainty-aware decision-making.

Despite the strong overall performance, several limitations should be acknowledged. At distances beyond 15m, the Ouster OS1-32 returns fewer than five points per cone cluster, increasing the likelihood of missed detections or poorly conditioned feature estimates. Furthermore, the system has not been evaluated under adverse weather conditions such as rain or fog.

In summary, we presented a lightweight LiDAR-based cone detection pipeline achieving 98.33% F1-score and 3.13 ms CPU-only runtime. Beyond FSD racing, the approach is applicable to road environments and industrial settings where cones define navigation boundaries. Future work will incorporate color detection and robustness evaluation under adverse weather conditions.

## Acknowledgements

The authors gratefully acknowledge the BME Formula Racing Team members for their contribution to vehicle development, data collection, and integration, and thank Dr. László Bokor and Prof. Tamás Szirányi for their help and support. Furthermore, the authors acknowledge Bosch for supporting the publication of this work. The authors indicate that generative AI tools were used for language polishing and grammar improvement.